\documentclass[preprint,12pt]{elsarticle}

\usepackage{graphicx}
\usepackage{subfig}

\usepackage{amsmath}

\usepackage[colorlinks,linkcolor=blue]{hyperref}

\usepackage[top=2cm, bottom=2cm, left=2cm, right=2cm]{geometry}  

\makeatletter  
\newif\if@restonecol  
\makeatother

\usepackage[ruled,vlined,lined]{algorithm2e}
\usepackage{algpseudocode}

\usepackage{booktabs}
\usepackage{multirow}

\usepackage{lineno}

\begin{document}
\begin{frontmatter}


\title{A Semantic Consistency Feature Alignment Object Detection Model Based on Mixed-Class Distribution Metrics}
\author[mymainaddress]{Lijun Gou}
\ead{getglj@hust.edu.cn}
\author[mymainaddress]{Jinrong Yang\corref{mycorrespondingauthor}}
\ead{yangjinrong@hust.edu.cn}
\author[mymainaddress]{Hangcheng Yu}
\author[mymainaddress]{Pan Wang}
\author[mymainaddress]{Xiaoping Li}
\cortext[mycorrespondingauthor]{Corresponding author}
\author[mymainaddress]{Chao Deng}
\address[mymainaddress]{State Key Laboratory of Digital Manufacturing Equipment and Technology, Huazhong University of Science and Technology, Wuhan, 430074, China.}

\begin{abstract}
 Unsupervised domain adaptation is critical in various computer vision tasks, such as object detection, instance segmentation, etc. They attempt to reduce domain bias-induced performance degradation while also promoting model application speed. Previous works in domain adaptation object detection attempt to align image-level and instance-level shifts to eventually minimize the domain discrepancy, but they may align single-class features to mixed-class features in image-level domain adaptation because each image in the object detection task may be more than one class and object. In order to achieve single-class with single-class alignment and mixed-class with mixed-class alignment, we treat the mixed-class of the feature as a new class and propose a mixed-classes $H-divergence$ for object detection to achieve homogenous feature alignment and reduce negative transfer. Then, a Semantic Consistency Feature Alignment Model (SCFAM) based on mixed-classes $H-divergence$ was also presented. To improve single-class and mixed-class semantic information and accomplish semantic separation, the SCFAM model proposes Semantic Prediction Models (SPM) and Semantic Bridging Components (SBC). And the weight of the pix domain discriminator loss is then changed based on the SPM result to reduce sample imbalance. Extensive unsupervised domain adaption experiments on widely used datasets illustrate our proposed approach's robust object detection in domain bias settings. 
\end{abstract}
\end{frontmatter}
\section{Introduction}
\label{S:1}
By virtue of large-scale of datasets like ImageNet~\cite{imagenet}, Open Image~\cite{openimage}, and MS COCO~\cite{coco}, Deep Convolutional Neural Networks (DCNN) have promoted object detection task to achieve remarkable progress. Object detection task requires a strong understanding for a specific scene including texture, brightness, viewing angle etc. This strong demand is difficult to be satisfied when applying a detector to different scenarios, causing performance degradation. This phenomenon is the notorious domain shift or dataset bias issues. While collecting and annotating data for related scenarios is capable of tackling this issue, it suffers enormous resources to achieve the expected results.

\begin{figure}[]
  \centering
  \subfloat[]{\label{cla_task}
  \includegraphics[width=0.45\linewidth]{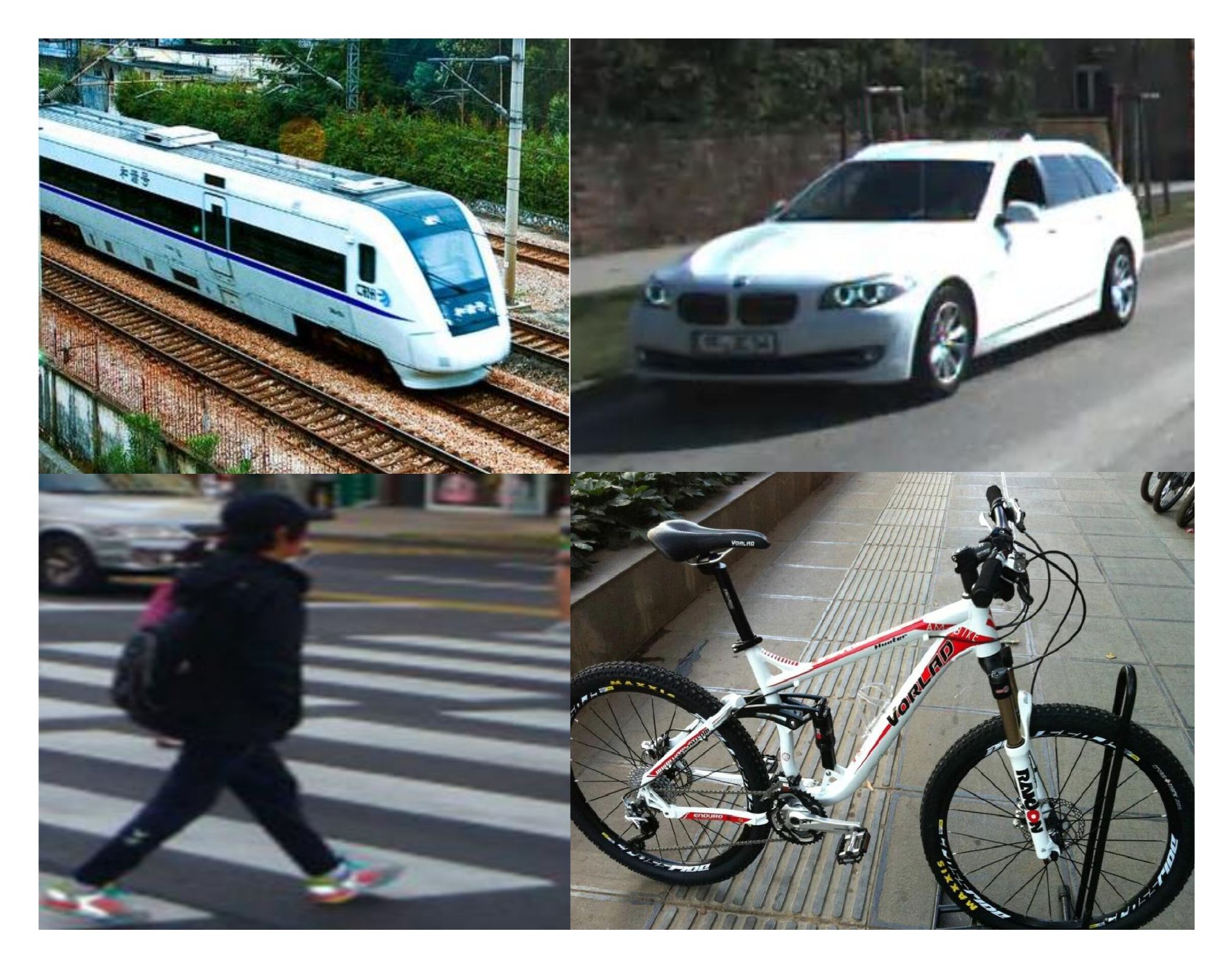}}
  \subfloat[]{\label{dete_task}
  \includegraphics[width=0.428\linewidth]{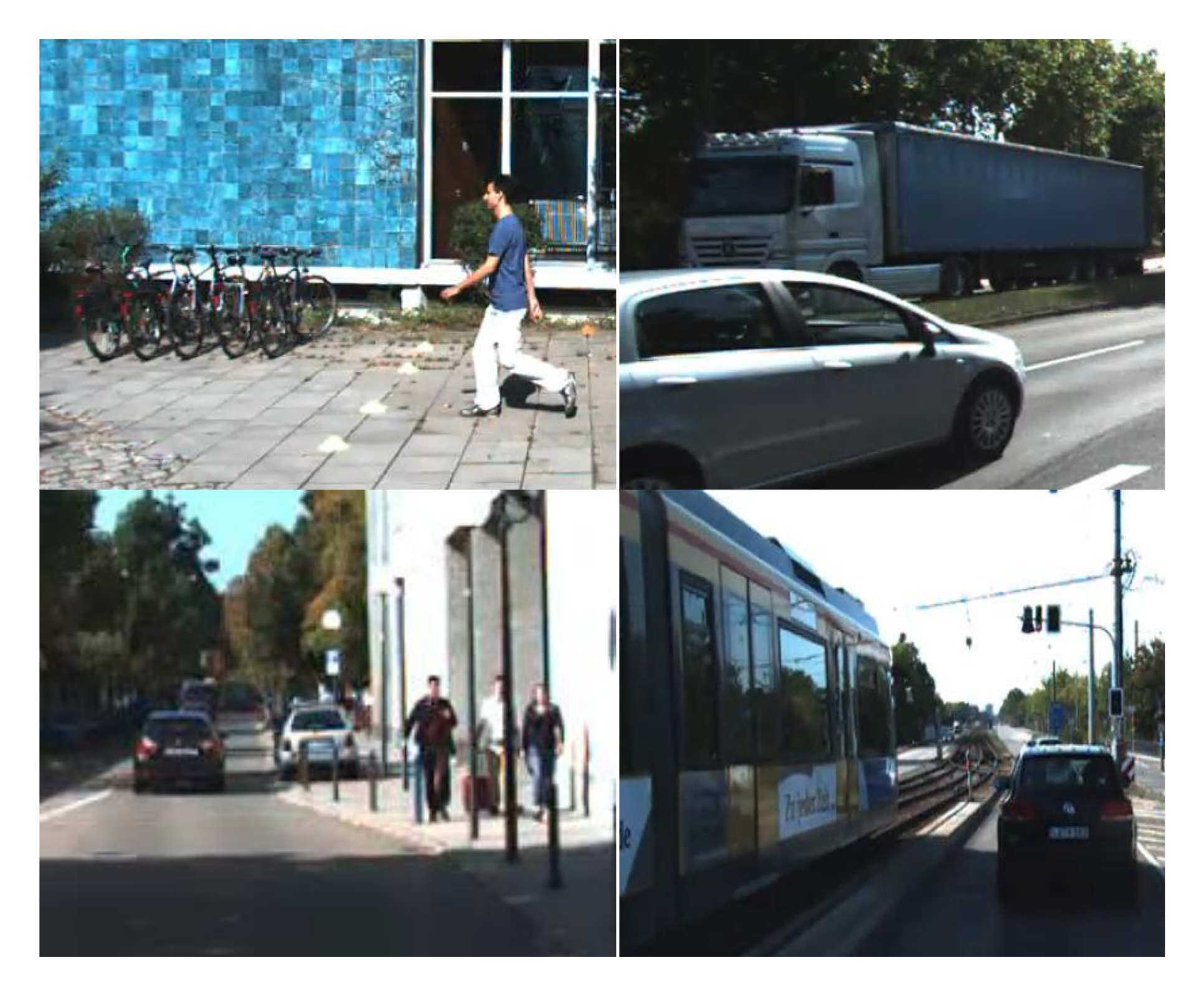}}
  \caption{(a) is a classification task ; (b) is a object detection task. The "person", "bike", "train" and "car" are the common classes on the classification and object detection task. As shown in Figure (a), there are only one class on each image. On the other hand, there are more than one classes on object detection task, shown in Figure (b).}
  \label{task}
  \end{figure}

To eliminates this barrier, Unsupervised Domain Adaptation approach (UDA) is tailored to address this problem without extra annotated data, which aims at adaptively conveying knowledge from the source domain to the target domain. Several advanced methods~\cite{chen2018domain,saito2019strong,he2019multi,xu2020exploring} employ the domain classifier to distinguish the discrimination and synergistically train a feature extractor by using the Gradient Reversal Layer (GRL), which aims at learning a consistent representation for target domain. These methods are based on the literature in \cite{ganin2016domain}, which aims to estimate the $H-divergence$ \cite{ben2010theory} between the source domain and target domain. We point out that these works~\cite{chen2018domain,saito2019strong,he2019multi,xu2020exploring} only focus on the $H-divergence$ measures for a single class on object detection task. However, object detection task is often required to locate and classify multiple categories of objects that may block each other, which is different from the classification task, as shown in Figure~\ref{task}. Due to this dilemma, the current methods \cite{chen2018domain,saito2019strong,he2019multi,xu2020exploring} are hard to transfer precise clue from source domain to target domain. In other word, the mixed semantic features are difficult to decoupled, causing to learn confusing information and failing to achieve the oracle performance.

\begin{figure}[]
\centering
\includegraphics[height=0.5\linewidth]{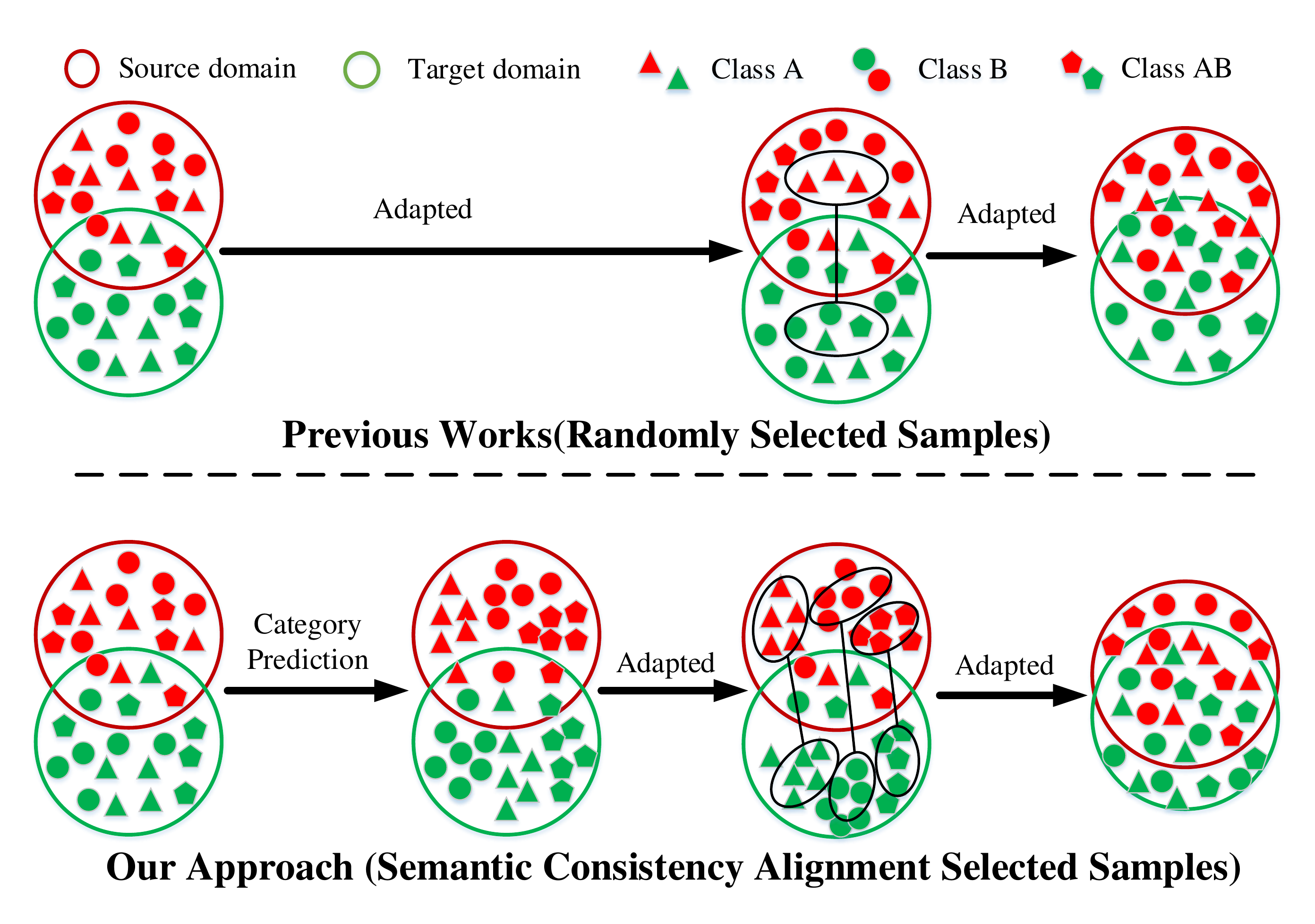}
\caption{As shown in the figure, previous works are based on the $H-divergence$ of single-class, which will cause the single-class to align with the mixed-class of features, such as the classes A and AB. Our works define a new mixed-class, such as the class AB, which is combined the feature of classes A and B. As we can see in the figure, if we calculate the mixed-classes $H-divergence$, it will reduce the error of totally $H-divergence$ and negative transfer.}
\label{contr_task}
\end{figure}

To address the multi-classes $H-divergence$ dilemma, we act the feature containing multiple categories as a new mixed category. As shown in Figure~\ref{contr_task}, given the original categories A and B which appears in the input image (or feature map), we add a new category named AB. Afterwards, we provide a Mixed-Classes $H-divergence$ ($MCH-divergence$) with the mixed classes to obtain a more accurate feature distribution. We also propose a semantic prediction module based on the $MCH-divergence$ to identity the mixed classes feature, as well as a semantic bridge component to increase semantic separability between single-class and mixed-class features. The two modules are used together to provide homogeneous feature alignment and reduce negative transmission.

On the other hand, according to the receptive field theory, the larger a feature's receptive field, the greater probability of a positive sample, and vice versa, the greater probability of a negative sample, which will result in a severe sample imbalance for the domain classification network in the domain adaptive object detection. Recent works~\cite{chen2018domain,saito2019strong,he2019multi,xu2020exploring} estimate the $H-divergence$ of each feature in the local feature map, which is different from the classification task. As shown the loss of domain adaptation(DA) below:
\begin{equation}
\begin{aligned}
   L_{DA}=\frac{1}{HW}\sum_{u,v}[\hat{D}logD^{(u,v)}+(1-\hat{D})log(1-D^{(u,v)})]
\end{aligned}
\label{domain-loss}
\end{equation} 
Here, $D^{(u,v)}$ is the output of the domain classification network, $\hat{D}$ is the label of the feature. $H$ and $W$ are the height and width of the feature map, this means the total number of features is $HXW$. There will be a large number of positive samples or negative samples for each iterate in training, which will cause the loss of domain adaptation imbalance. To address this issue, we presented a semantic attention domain adaptation loss based on the output of the semantic prediction module. Because the semantic prediction module's output contains the probability of the classes in the feature, we can derive the foreground attention map from it. As a result, we can use the foreground attention map to alter the domain classification network's loss to mitigate the consequences of sample imbalance.

The main contributions of this work are three-fold:
\begin{itemize}
\item[1)] We propose a Mixed-Classes $H-divergence$ ($MCH-divergence$), which achieves a more accurate feature distribution distance between source and target domain by measuring the $H-divergence$ with single and mixed classes. 
\item[2)]Based on $MCH-divergence$, we propose a domain adaption model for object detection with a semantic prediction module and a semantic bridge component that may strengthen feature semantic information and realize the semantic separation of a single class and a mixed class. Thus, they can achieve homogeneous feature alignment and reduce negative transfer.
\item[3)]We propose a semantic attention domain loss to reduce the sample imbalance problem, which is based on the output of semantic prediction module.
\end{itemize}

\section{Related Work}
\label{S:2}
\subsection{Metric Theory of Domain Adaptation}
\label{subS:2.1}
Inconsistent probability distributions between training and test samples may result in poor model performance on target samples. As a result, evaluating the difference in probability distribution between the source and target domains becomes crucial. Daniel Kifer \cite{A-distance} proposed an $A-distance$ to measure the probability distributions of the source domain and target domain. The $A-distance$ not only guarantees the reliability of observed changes statistically, but it also gives meaningful descriptions and quantification of these changes. To realize the $A-distance$, Ben-David \cite{H-distance} limited the complexity of the true classification function $f$ in terms of their hypothesis class $H$ and proposed an $H-divergence$. Based on the $H-divergence$, John Blitze \cite{H2H-distance,ben2010theory} assumed that there are some asymmetric classification functions in terms of their hypothesis class $H$ and proposed an $H\Delta H-divergence$. The $H\Delta H-divergence$ distance proves that the distribution distance between the source domain and target domain can be estimated by finite samples. The above distribution distance is assumed to be binary classification task, thus, Wang~\cite{wang2018stratified} proposed a multi-class maximum mean discrepancy for the multi-class classification task. Currently, the distribution distances across domains are used for classification tasks; if we use these distribution distances to design object detection models, the mixed-class distribution distances will be ignored. Furthermore, the models will cause single-class alignment features to be converted to mixed-class features.
   
\subsection{Domain Adaption for Object Detection}
\label{subS:2.2} 
  The domain adaptive object detection adapts to the target data through the existing annotations in the source domain to solve the problem of domain biases. Yaroslav Ganin et al.~\cite{ganin2016domain} first proposed GRL layers and domain prediction networks to reduce domain bias for classification tasks. Then Yuhua Chen et al.~\cite{chen2018domain} first used GRL to reduce the domain bias for object detection tasks. After that, there are many feature distribution matching methods to reduce domain bias, which are designed based on $H\Delta H-divergence$, such as the Multi-adversarial method \cite{he2019multi,xie2019multi}, which are proposed for Multi-layer feature alignment of the backbone for object detection, Strong-weak distribution alignment methods \cite{saito2019strong,xu2020exploring}, Memory Guided Attention for Category-Aware method \cite{vibashan2021mega} etc. And there are also many other methods to reduce the domain bias for object detection, such as image style transformation \cite{hsu2020progressive,toldo2020unsupervised,gou2022carton}, pseudo label technology \cite{li2021category,khodabandeh2019robust} etc. And our method is designed based on the Strong-weak distribution alignment methods~\cite{saito2019strong} in this paper. 

  \subsection{Attention Mechanism}
\label{subS:2.3} 
The visual attention mechanism is a one-of-a-kind brain signal processing mechanism for human eyesight as well as a resource allocation mechanism. The attention mechanism in a neural network can be thought of as a resource allocation system that redistributes resources based on the relevance of objects. There are many methods to achieve attention mechanisms such as the K-mean methods \cite{zhu2019adapting}, Attention-based Region Transfer \cite{zheng2020cross}, Local–Global Attentive Adaptation \cite{zhang2021local} etc. Different domain adaptation attention mechanisms are proposed in this paper for the features of different layers to allow the model to pay greater attention to a few examples.

\section{Domain Divergence for Object Detection}
\label{S:3}
In this paper, we denote the labeled source domain dataset as $S$ and its distributions as $D_S$, unlabeled target dataset as $T$ and the target domain distributions is $D_T$. $S=\{(x_i^s,y_i^s)\}_{i=1}^{n_s}\sim D_S$ contains the training image $x_i^s$ and its corresponding object bounding box label $y_i^s$ for $i=1,2,...,n_s$. For each bounding box label $y_i^s=\{y_c,y_b\}$, $y_c$ is the category and $y_b$ is the coordinates of the corresponding bounding box. $T=\{x_i^t\}_{i=1}^{n_t}\sim D_T$ only contains the obtained training image $x_i^t$. The goal of an unsupervised domain adaptation learning algorithm is to build a Detector~(denoted by $h$) with a low target risk on target domain:
\begin{equation}
\label{risk_eup}
\varepsilon_{T} =\underset{(x,y)\sim D_T}{P_r}(h(x)\ne y)
\end{equation}
Where $x$ is the input image from the target domain, $y$ is the true bounding box label of the objects in $x$.

Because only the source domain has labeled data for domain adaptation tasks, many approaches restrict the target error by the sum of the source error and a measure of distance between the source and target distributions. When the distributions of the source and target risks are similar, these methods assume that the source risk is a good predictor of the target risk. Several notions of distance have been proposed for domain adaptation,such as $H$-divergence~\cite{ben2010theory}, Renyi divergence~\cite{mansour2012multiple} etc. In this paper, we focus on the $H$-divergence. Let us denote by $x$ a feature vector, we also denote by $\eta:x \rightarrow\{0, 1\}$ a domain classifier, which aims to predict the source samples $x_S$ to be $0$, and target domain sample $x_T$ to be $1$. Suppose $H$ is the set of possible domain classifiers, the $H$-divergence defines the distance between two domains as follows~\cite{ganin2016domain,chen2018domain}:
\begin{equation}
\begin{aligned}
d_{H}(S,T) &= 2\underset{h\in EDT}{sup}|Pr_{D_S}(I(h))-Pr_{D_T}(I(h))|\\
           &= 2[1-\underset{\eta\in H}{\min}(err_S(\eta)+err_T(\eta))]
\end{aligned}
\label{H_distance}
\end{equation}
Where $DET$ is the possible detector, $I(h)$ is a indicator function which is 1 if predicate a is true, and 0 otherwise, $I(h)=\{x\in Z:h(x)=y,h\in DET\}$. And $Z$ is the feature of objects. Simultaneously, $err_S$ and $err_T$ represent the prediction errors of $\eta(x)$ on source and target domain samples, respectively. Furthermore, inspired by Wang~\cite{wang2018stratified}, if there is more than one class which are denoted by $C=\{1,\dots,K\}$ and only one class in each image, the $H$-divergence will be written as follow:
\begin{equation}
\begin{aligned}
d_{H}(S,T) &=\sum_{c=1}^{K}d_H^c(S,T)\\
           &= \sum_{c=1}^{K}[1-\underset{\eta\in H_c}{\min}(err_S(\eta)+err_T(\eta))]
\end{aligned}
\label{H_C_distance}
\end{equation}

As we all know, there is only one class in each image for the classifier tasks, but in object detection tasks, there is more than one class in each image shown in Figure~\ref{task}. According to the receptive field theory, each feature in the backbone may have many categories in object detection tasks. As a result, the recent works~\cite{wang2018stratified,ben2010theory,mansour2012multiple} indicated by Eq.~\ref{H_C_distance} cannot accurately reflect the distribution difference between the source domain and the target domain in object detection. Assuming that distinct combination features of the classes represent a new class, we may obtain the Mixed Class $H$-divergence($MCH$-divergence) with source and target domains in the object recognition task as follows:
\begin{equation}
\begin{aligned}
d_{MCH}(S,T) &=\sum_{c_1=1}^{K}d_H^{c_1}(S,T)+ \sum_{c_1=1}^{K}\sum_{c_2=1}^{K}d_H^{c_1\ne c_2}(S,T)\\
           &+\dots+\sum_{c_1=1}^{K}\sum_{\dots}^{K}\sum_{c_K=1}^{K}d_H^{c_1\ne c_2\ne\dots\ne c_K}(S,T)\\
\end{aligned}
\label{H_object}
\end{equation}
Using this $MCH$-divergence notion, one can obtain a probabilistic bound~\cite{ganin2016domain} about detector risk on target domain:
\begin{equation}
\label{err_bound}
\varepsilon_T(h)\leq\varepsilon_S(h)+\frac{1}{2}d_{MCH}(S,T)+ \lambda
\end{equation}
 As shown in the Eq.\ref{err_bound}, the source domain error $\varepsilon_{S}(h)$, the distribution distance $d_{MCH}(S,T)$ calculated between the source domain and the target domain, and a constant $\lambda$ make up the error supremum $\varepsilon_{T}(h)$ of the target domain.

\section{Method}
\label{S:4}
\begin{figure}[]
\centering
\includegraphics[width=1.0\linewidth]{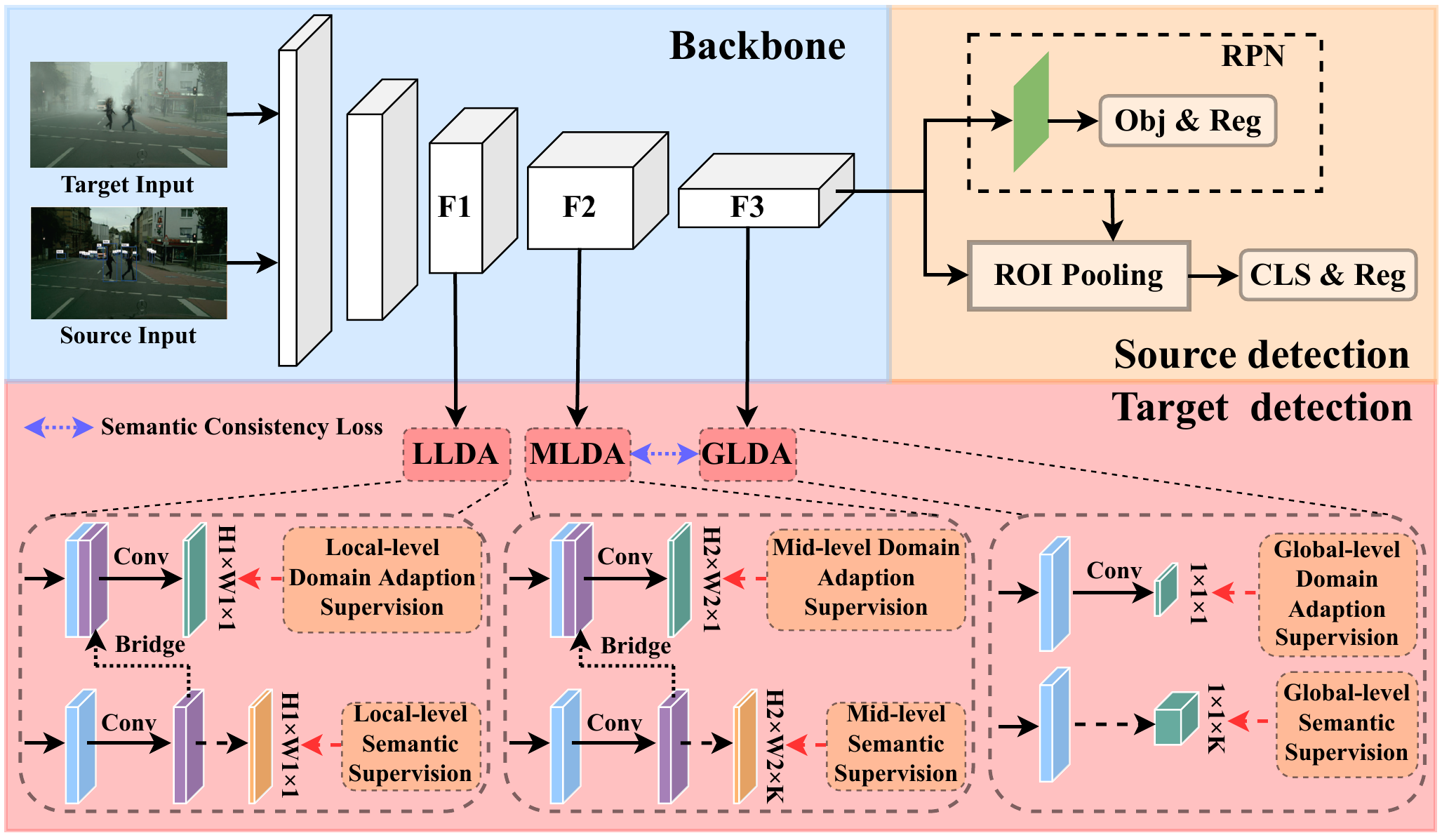}
\caption{The core framework of the semantic consistency domain adaptive object detection model. LLDA, MLDA, and GLDA indicate local level domain adaptive module, middle level domain adaptive module, and global level domain adaptive module, respectively.}
\label{SC-SW-DA}
\end{figure}

\subsection{Framework Overview}
\label{subS:4.1} 
 Figure~\ref{SC-SW-DA} depicts an overview of our semantic consistency framework. The main idea behind this framework is to achieve $MCH-divergence$ on the domain adaptation object detection task based on the SW-faster model \cite{saito2019strong}. To do this, we design an domain adaptive object detection model capable of adapting alignment semantics and corresponding features from fine-grain to coarse-grain. The over pipeline is show in Figure~\ref{SC-SW-DA}, we design three level of semantic prediction modules including local, middle and global granularities, which provides an implicit spatial alignment effect and yield an efficiently semantic clue to bridge the gradient reverse layer (GRL). Furthermore, a semantic consistency loss is tailed to keep multi-level semantic alignment.
\subsection{Semantic Prediction Modules}
\label{subS:4.2}
\begin{figure}[]
\centering
\includegraphics[width=0.65\linewidth]{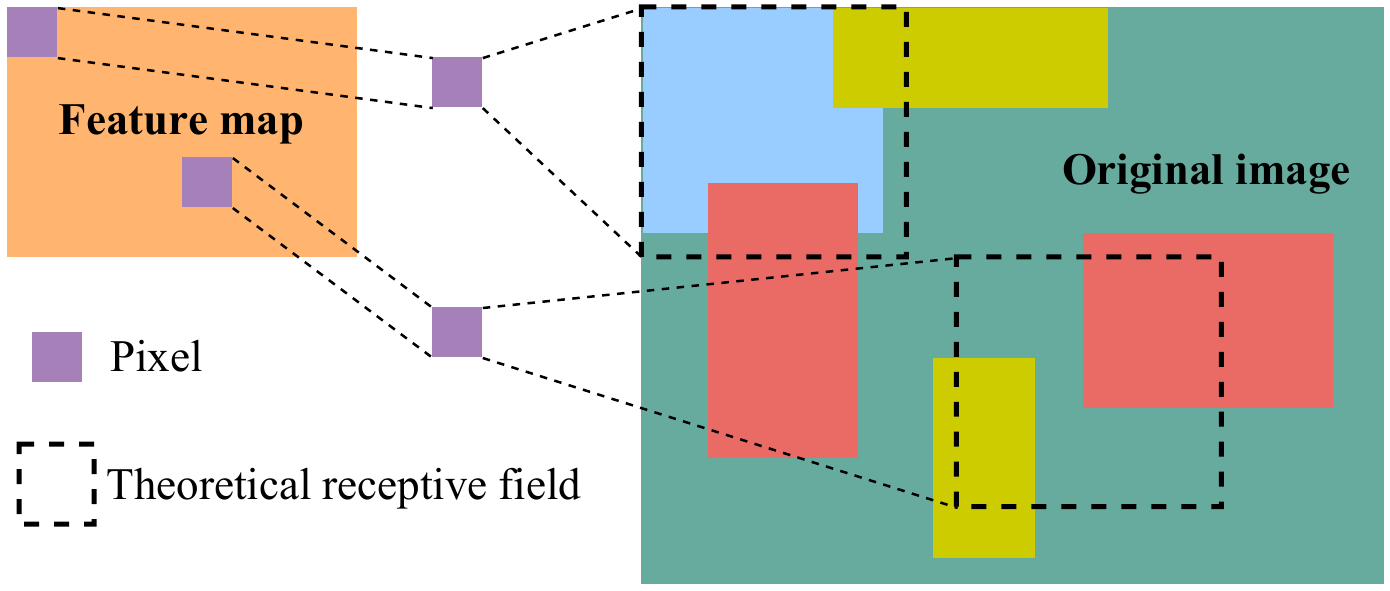}
\caption{The example of the feature receptive field. In the original image, the rectangles of different colors represent different categories and the black rectangle is the receptive field.}
\label{semantic_information}
\end{figure}

 As we all know, the deeper the backbone net, the larger the receptive field size of the feature. The feature's receptive field size is an area of the original image(shown in Figure~\ref{semantic_information}), and the size of the receptive field can be calculated using Eq.~\ref{field_eq}. Here, the $l_k$ is the receptive field size of kth convolution layer, $f_k$ is the size of kth convolution kernel, $S_i$ is the stride of the ith convolution kernel.
 \begin{equation}
\begin{aligned}
l_k=l_{k-1}+[(f_k-1)*\prod_{i=1}^{k-1}S_i]
\end{aligned}
\label{field_eq}
\end{equation}

There is more than one class in the receptive field, as shown in Figure~\ref{semantic_information}, and it can get the category types contained in each feature based on the receptive field size. If there are $K$ categories, it can create a $K$-dimensional semantic vector for each feature represented by $S_{vector}$. Furthermore, each element in $S_{vector}$ represents a single type of category, which can be calculated by calculating the overlap area between the receptive field and the ground truth. Assume the receptive field size is $W\times W$, the area is denoted by $S_w$, the area of ground truth is denoted by $S_g$, and the overlap area is denoted by $S_{w\cap g}$. Then we use Eq.~\ref{category_exists} exists to see if the category exists in the receptive field; if it does, the value is $1$; otherwise, it is $0$.
 \begin{equation}
\begin{aligned}
\frac{S_{g\cap w}}{S_p} \geq \zeta
\end{aligned}
\label{category_exists}
\end{equation}
 Where $zeta$ is a hyperparameter determining whether or not the class exists. There are other examples, such as too large receptive fields or too small object bounding boxes. As a result, an expected area named $S_p$, which is the small area between $S_w$ and $S_g$, is used. The semantic vector labeling algorithm is illustrated in Algorithm.\ref{S_V_L_A}.
   \begin{algorithm}[]
    \caption{Semantic Vector Labeling Algorithm Based on Receptive Field}  
    \LinesNumbered 
    \label{S_V_L_A}  
    \KwIn{ The ground truth boxes:\quad$gt\_boxes$\\
           \quad\quad\quad\quad The category vector of the ground truth boxes:\quad$gt\_classes$\\
           \quad\quad\quad\quad The number of classes:\quad$num\_class$\\
           \quad\quad\quad\quad Hyperparameter:\quad $\zeta$\\
           \quad\quad\quad\quad The size of receptive field:\quad $W \times W$} 
    \KwOut{the semantic vector of feature:\quad $S\_{vector}$}  
    
    \textbf{Initialize:} {$S\_vector=\{0,0,...,0\}$\\
    the number of ground truth:\quad $num\_gt \leftarrow len(gt\_boxes)$\\
    the area of receptive field:\quad $S_w\leftarrow area(W)$\\}
    \For{ $id\rightarrow num\_gt$ }
    {
     $S_g\leftarrow area(gt\_boxes[id])$\\
      $S_p\leftarrow min(S_w,S_g)$\\
      Calculate the intersection area with $gt\_boxes[id]$ and $W$: $S_{{gt\_boxes[id]}\cap w}$\\
      \If{$\frac{S_{{gt\_boxes[id]}\cap W}}{S_p} \geq \zeta$}
       {
        Get the class index of the $gt\_classes[id]$ in  $S\_vector$: $index$\\
        $S\_vector[index]\leftarrow 1$
        }
    }
    return $S\_vector$
  \end{algorithm} 
  
 As we all know, the lower features of the backbone have more common structures, and the higher features of the backbone have more semantic features~\cite{yosinski2014transferable}. So three semantic prediction modules are proposed in this paper to get the semantic vector of the feature, as shown in Figure~\ref{SC-SW-DA}.

 \textbf{Local-level semantic prediction:} The receptive field of the lower feature in the model's backbone is small, as shown in $F1$ in Figure~\ref{SC-SW-DA}. As a result, there are a large number of features that are a part of the overall object. Thus, the local-level semantic prediction module only predicts whether a feature is in the foreground or background; it is a binary classification, and the binary classification loss is below:
\begin{equation}
\begin{aligned}
L_{S_l}=\frac{1}{H_1W_1}\sum_{i=0}^{H_1}\sum_{j=0}^{W_1}[\hat{y}_{i,j}log(y_{i,j})+(1-\hat{y}_{i,j})log(1-y_{i,j})]
\end{aligned}
\label{local-sem}
\end{equation}
where the $H_1,W_1$ is the size of feature map $F1$, $\hat{y}_{i,j}$ is the ground truth class which is located at $i,j$ in $F1$. $y_{i,j}$ is the output of local-level semantic prediction module.
 
\textbf{Mid-level semantic prediction:} The receptive field of the middle feature in the model's backbone is medium, as shown in $F2$ in Figure~\ref{SC-SW-DA}. There are many features in $F2$ that contain the entire object based on the size of the receptive fields. As a result, the mid-level semantic prediction module is proposed to predict the semantic vector of each feature; it is a multi-classification task with a classification loss as below:
\begin{equation}
\begin{aligned}
L_{S_m}=\frac{1}{H_2W_2}\sum_{i=0}^{H_2}\sum_{j=0}^{W_2}\sum_{c=1}^{K}[\hat{y}^c_{i,j}log(y^c_{i,j})+(1-\hat{y}^c_{i,j})log(1-y^c_{i,j})]
\end{aligned}
\label{mid-sem}
\end{equation}
where the $H_2,W_2$ is the size of feature map $F2$, $\hat{y}^c_{i,j}$ is the ground truth class which is located at $i,j$ in $F2$. $y^c_{i,j}$ is the output of mid-level semantic prediction module.
 
\textbf{Global-level semantic prediction:} The receptive field of the global feature in the model's backbone is too large, as shown in $F3$ in Figure~\ref{SC-SW-DA}. And the semantic information is self-evident. As a result, we only predict the class semantic information in the entire image, and the classification loss is below:: 
\begin{equation}
\begin{aligned}
L_{S_g}=\sum_{c=1}^{K}[\hat{y}^clog(y^c)+(1-\hat{y}^c)log(1-y^c)]
\end{aligned}
\label{global-sem}
\end{equation}
where the $K$ is number of classes in source domain and target domain, $\hat{y}^c$ is the ground truth class, $y^c$ is the output of global-level semantic prediction module.

\subsection{Semantic Consistency Domain Adaptive Modules}
\label{subS:4.3}

 \begin{figure}[]
\centering
\includegraphics[width=0.8\linewidth]{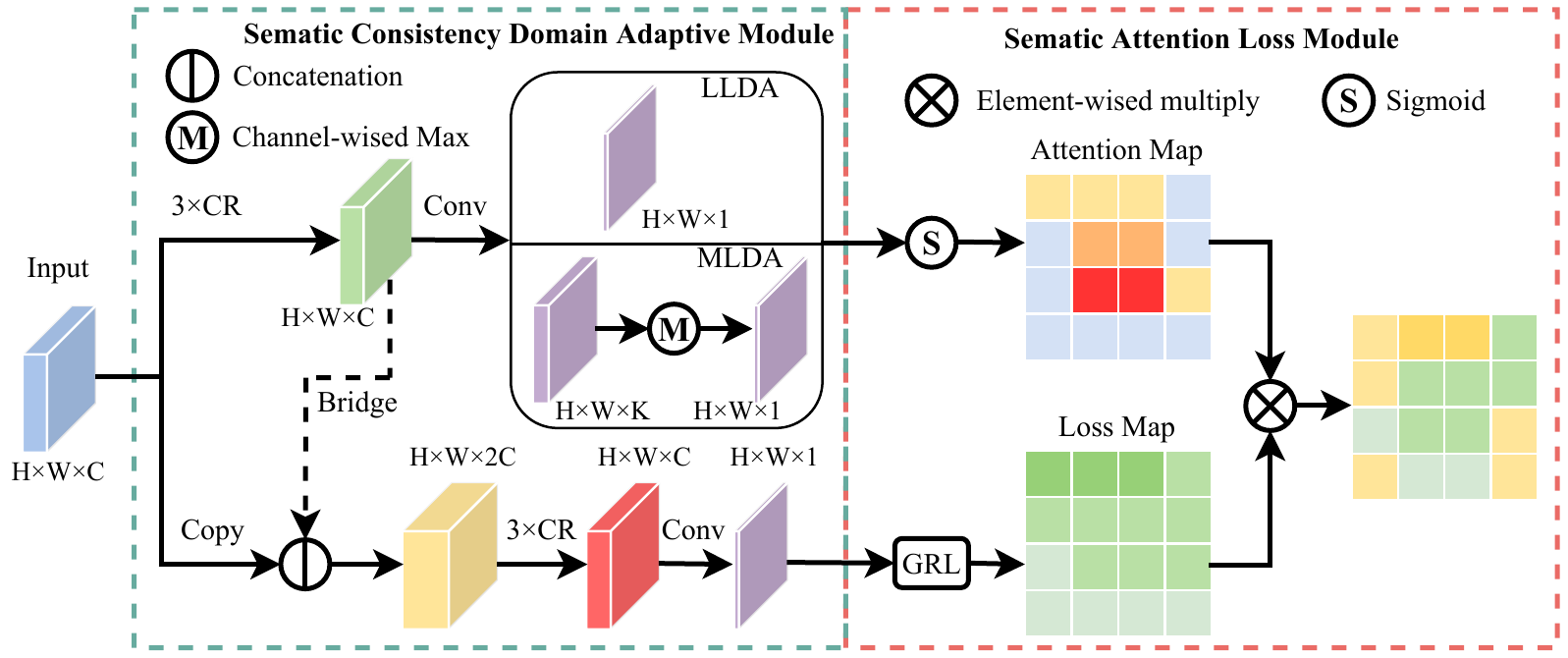}
\caption{Semantic consistency domain adaptive module with semantic attention map. N$\times$CR means conducting N modules of 1$\times$1 convolution with ReLU. The solid line box means local level domain adaptive module directly uses the convolution layers output as Semantic Attention Map, while middle level domain adaptive module employs a convolution layer with channel-wised max pooling to a Semantic Attention Map.}
\label{SC-A-L}
\end{figure}

There are three domain adaptive modules for features (such as $F1,F2,F3$) in backbone, as illustrated in Figure~\ref{SC-SW-DA}. The pixel-level feature alignments in the local-level domain adaptation and mid-level domain adaptation models are the same as the local adaptation module in SW-faster \cite{saito2019strong}. However, there is insufficient semantic information in the local or middle convolution layers ($F1, F2$) to distinguish between mixed-class and single-class features, causing the single-class to align with the mixed-class. To achieve homogeneous feature alignment, semantic bridge components are proposed. And because the global feature has more semantic information, the global-level domain adaptation is the same as SW-faster.
  
  \textbf{Semantic Bridge Components:} In the object detection task, the feature may contain more than one category, according to the theory of receptive field. Furthermore, in the current work, the domain adaptive prediction module may cause the single-class to align with the mixed-class. So, we proposed a $MCH-divergence$ to measure the distance between the source and target domains in order to reduce negative transfer. In the domain adaptation module, we proposed a semantic bridge component to improve the semantics of the features depicted by the green line in Figure~\ref{SC-A-L}, which concatenates the penultimate layer features of the semantic prediction module into the corresponding feature maps in the backbone of models channel by channel. The semantic bridge component has two advantages: 1)It will improve the semantic separability of single-class and mixed-class features in the semantic space, as well as achieve feature alignment within the same class; 2)It will implement adaptive learning of semantic prediction modules.
  
  \textbf{Local-level Domain Adaptation:} The $F1$ feature in the backbone (as shown in \ref{SC-SW-DA}) is used for local-level domain adaptation, and the $F1$ channel is $C1$. Assuming that the local-level semantic feature channel is $S1$, the feature channel of the local-level domain adaptation inputting is $C1+S1$. And the goal function is as follows::
  \begin{equation}
  \begin{aligned}
   L_{l}=\frac{1}{H_1W_1}\sum_{u,v}[\hat{D}logD_l^{(u,v)}+(1-\hat{D})log(1-D_l^{(u,v)})]
  \end{aligned}
  \label{DA-local}
  \end{equation}
  Where, the $\hat{D}$ is the domain label, source domain is 0, target domain is 1. And the $D_l^{(u,v)}$ is the output of local-level domain adaptation module at $u,v$ position. 
  
  \textbf{Mid-level Domain Adaptation:} The $F2$ backbone feature is used for mid-level domain adaptation, and the $F2$ channel is $C2$. Assuming that the mid-level semantic feature channel is $S2$, the feature channel of the mid-level domain adaptation inputting is $C2+S2$. And the goal function is as follows::
    \begin{equation}
  \begin{aligned}
   L_{m}=\frac{1}{H_2W_2}\sum_{u,v}[\hat{D}logD_m^{(u,v)}+(1-\hat{D})log(1-D_m^{(u,v)})]
  \end{aligned}
  \label{DA-mid}
  \end{equation}
  Where, the $D_m^{(u,v)}$ is the output of mid-level domain adaptation module at $u,v$ position. 
  
  \textbf{Global-level Domain Adaptation:} Since the $F3$ feature layer has enough semantic information, the domain adaptation module has no semantic bridge component and its domain adaptation module is consistent with SW-faster~\cite{saito2019strong}. And the goal function is as follows:
      \begin{equation}
  \begin{aligned}
   L_{g}=\hat{D}(1-D_g)^\gamma logD_g+(1-\hat{D})D_g^\gamma log(1-D_g)
  \end{aligned}
  \label{DA-global}
  \end{equation}
  Where, $\gamma$ is the hyperparameters from FocalLoss~\cite{lin2017focal}, the $D_g$ is the output of global-level domain adaptation module.

\begin{figure}[t]
\centering
\includegraphics[width=0.7\linewidth]{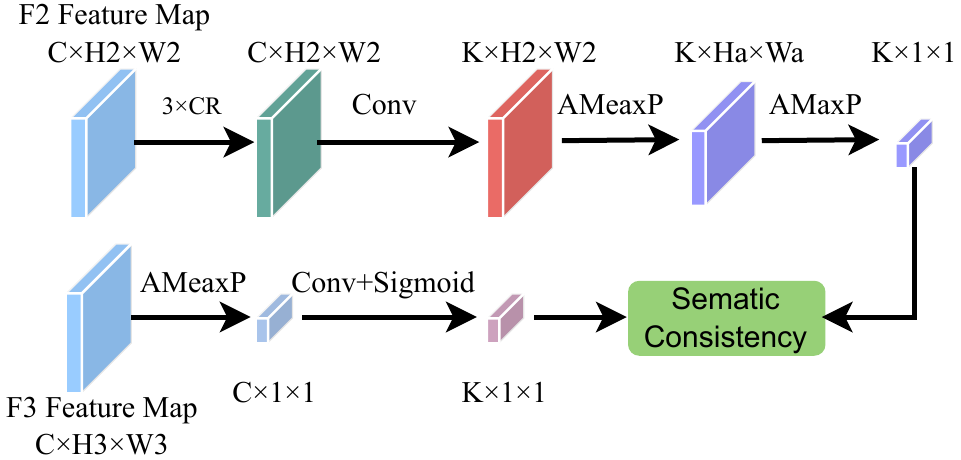}
\caption{Semantic consistency regularization for source data. N$\times$CR means conducting N modules of 1$\times$1 convolution with ReLU. AMeanP and AMaxP means adaptive mean pooling and adaptive max pooling, respectively. FC means using a fully connection layer.}
\label{SC-R-L}
\end{figure}

\subsection{Semantic-based Attention Maps}
\label{subS:4.4}
Since the local-level and mid-level domain adaptation use pixel features for a feature adaptation, the positive and negative samples are very unbalanced. Therefore, we use the output of the semantic prediction module as semantic attention maps to adjust the loss of small samples in the local and middle domain adaptive model.

\textbf{Local semantic attention map:}
Because the receptive field of the feature in $F1$ is small, the number of negative samples(such as backgrounds) is much larger than the number of positive samples(such as objects). And the output of the local-level semantic prediction module is the probability of the foreground target, which can be directly used to adjust the local-level domain loss weight of positive samples, as shown in Figure~\ref{SC-A-L}, the $K$ is 1. And the loss function with attention map of the local-level domain adaptation is as follows:
  \begin{equation}
  \begin{aligned}
   \hat{L}_{l}=(1+P_l)\otimes L_l=\frac{1}{H_1W_1}\sum_{u,v}(1+P_l^{(u,v)})[\hat{D}logD_l^{(u,v)}+(1-\hat{D})log(1-D_l^{(u,v)})]
  \end{aligned}
  \label{DA-local-new}
  \end{equation}
  Where, $P_l$ is the output of local-level semantic prediction module, the $\otimes$ is the position-wise multiplication.
  
  \textbf{Middle semantic attention map:} There are more positive samples because the $F2$ has a larger receptive field. As a result, we should increase the number of negative samples. Because the output of the mid-level semantic prediction module contains more than one class, the probability of foreground object is represented by the maximum value (represented by $P_m$) of the output for the $K$ category, as shown in Figure~\ref{SC-A-L}. As a result, the mid-level domain adaptation goal function with attention map is as follows: 
  \begin{equation}
  \begin{aligned}
   \hat{L}_{m}=(1+1-P_m)\otimes L_m=\frac{1}{H_2W_2}\sum_{u,v}(2-P_m^{(u,v)})[\hat{D}logD_m^{(u,v)}+(1-\hat{D})log(1-D_m^{(u,v)})]
  \end{aligned}
  \label{DA-mid-new}
  \end{equation}

\subsection{Semantic Consistency Regularization and Total Loss}
\label{subS:4.5}
\textbf{Semantic Consistency Regularization:} Theoretically,  the categories type of the mid-level semantic prediction output is the same as the global-level prediction. As a result, the categories in $F2$ should be the same as those in $F3$. At the same time, because the target domain lacks labeling information, this paper focuses solely on the semantic consistency of the source domain.

The output of the mid-level semantic prediction module is a probability map of all categories, as shown in Figure~\ref{SC-R-L}. The output of the global-level semantic prediction module, on the other hand, is a vector of the $K$ dimension. In the mid-level semantic output, there are $H 2*W 2$ values for each class; if the maximum probability for each class is directly calculated, the singular values will dominate the results. To reduce the influence of singular values, adaptive mean pooling is used first, and its output is $K*H a*W a$, followed by adaptive max pooling to get the categories, which will be the same as the output of the global-level semantic prediction module. And then there's the loss of semantic consistency:
  \begin{equation}
  \begin{aligned}
  L_{CR}=\sum_{c=1}^{K}[y_g^clog(y_m^c)+(1-y_g^c)log(1-y_m^c)]
  \end{aligned}
  \label{DA-L-L}
  \end{equation}
Where, the $y_g^c$ is the probability of the class $c$ in global-level semantic prediction output. The $y_m^c$ is the probability of the class $c$ after adaptive mean pooling and adaptive max-pooling based on the mid-level semantic prediction output in Figure~\ref{SC-R-L}.

\textbf{Total Loss:} Because the model in this paper is made up of three main components: semantic prediction modules, domain adaptation modules, and a Faster R-CNN detection network. When all of the modules form a multi-task learning system during model training, the total loss is as follows:
  \begin{equation}
  \begin{aligned}
  L_{all}=L_{det}+\lambda_{1}(\hat{L}_{l}+\hat{L}_{m}+L_g)+\lambda_{2}(L_{S_l}+L_{S_m}+L_{S_g})+\lambda_{3}L_{CR}
  \end{aligned}
  \label{all-loss}
  \end{equation}
 where $L_{det}$ is the loss of Faster-RCNN~\cite{ren2016faster} including softmax loss function and smooth l1 loss. $\lambda_{1}$, $\lambda_{2}$ and $\lambda_{3}$ are a trade-off coefficient to balance the Faster RCNN model with our components. Here, $\lambda_{1}=-1$, $\lambda_{2}=1$ and $\lambda_{3}=1$.

\section{Experiments}
\label{S:5}

\subsection{Experiment Setup}
\label{S:5.1}
\textbf{Datasets.} For domain adaptation experiments, six public datasets are used: Cityscapes \cite{cityscapes}, Foggy Cityscapes \cite{foggy-cityscapes}, BDD100k \cite{bdd100k} for similar domain adaptation experiments, and PASCAL VOC \cite{pascalvoc}, Clipart1k \cite{clipart1k}, WaterColor2K \cite{clipart1k} for dissimilar domain adaptation experiments.
\begin{itemize}
    \item \textbf{Cityscapes} focuses on capturing images of outdoor street scenes in common weather conditions from various cities. It includes 2,975 training images and 500 validation images taken during clear weather.
    \item \textbf{Foggy Cityscapes} are rendered from Cityscapes using depth maps. It stimulates the change of weather conditions, which is suitable to conduct foggy weather adaptation experiments. And the number of datasets is same with Cityscapes.
    \item \textbf{BDD100K} consists of 100k images, with 70k training images and 10k validation images annotated with bounding boxes. We mainly extract the images labeled as daytime as a sub-dataset for city scene adaptation experiments, which contains 36,728 training sets and 5,258 validation sets and the categories are same with the Cityscapes.
    \item \textbf{PASCAL VOC} contains 20 categories of common objects with bounding box annotations. Following~\cite{saito2019strong,xu2020exploring}, we employ PASCAL VOC 2007 and 2012 (16,551 images in total) for dissimilar domain adaptation experiments.
    \item \textbf{Clipart1k} contains 1k clipart images and belongs to the same category as PASCAL VOC~\cite{pascalvoc}. All 1k images will be used as target datasets for training and testing in this paper.
    \item \textbf{WaterColor2K} shares 6 categories with PASCAL and a total of 2K images. During training, 1K training images were used, and the model was evaluated using the remaining 1K test images.
\end{itemize}

\textbf{Implementation Details.} Following the default settings in \cite{saito2019strong,xu2020exploring}. all training and test images are resized such that the shorter side has a length of 600 pixels. By default, the backbone models for similar domain adaptation experiments are initialized using pre-trained weights of VGG-16 \cite{vgg16} on ImageNet, and for others, we follow the practices in \cite{saito2019strong,xu2020exploring} and use ResNet-101 \cite{he2016res101} as the detection backbone. We fine-tune the network with a learning rate of $10^{-3}$ for 50k iterations and then reduce the learning rate to $10^{-4}$ for another 20k iterations. The momentum of 0.9 and the weight decay of $5\times10^{-4}$ are used for VGG-16 based detectors, while for ResNet-101 based detectors, the weight decay is $1\times10^{-4}$. In all experiments, we employ RoIAlign \cite{he2017mask} for RoI feature extraction, $\zeta=0.6$ and the output of adaptive mean pooling is $10*10$.

\begin{table}[]
\centering
\caption{Weather Changes: The Cityscapes are the source datasets, and the Foggy Cityscapes are the target datasets. We present the mAP results for Foggy Cityscapes in eight categories and compare them to recent works.}
\begin{tabular}{l|ccccccccc}
\toprule[2pt]
                   & \multicolumn{1}{c}{person} & \multicolumn{1}{c}{rider} & \multicolumn{1}{c}{car}  & \multicolumn{1}{c}{truck} & \multicolumn{1}{c}{bus}  & \multicolumn{1}{c}{train} & \multicolumn{1}{c}{mcycle} & \multicolumn{1}{c}{bicycle} & \multicolumn{1}{c}{mAP}  \\ \hline
DA-faster~\cite{chen2018domain}  & \multicolumn{1}{c}{28.7}   & \multicolumn{1}{c}{36.5}  & \multicolumn{1}{c}{43.5} & \multicolumn{1}{c}{19.5}  & \multicolumn{1}{c}{33.1} & \multicolumn{1}{c}{12.6}  & \multicolumn{1}{c}{24.8}   & \multicolumn{1}{c}{29.1}    & \multicolumn{1}{c}{28.5} \\
Baseline:SW-faster~\cite{saito2019strong}  & \multicolumn{1}{c}{32.3}   & \multicolumn{1}{c}{42.2}  & \multicolumn{1}{c}{47.3} & \multicolumn{1}{c}{23.7}  & \multicolumn{1}{c}{41.3} & \multicolumn{1}{c}{27.8}  & \multicolumn{1}{c}{28.3}   & \multicolumn{1}{c}{35.4}    & \multicolumn{1}{c}{34.8} \\
SW-ICR-CCR~\cite{xu2020exploring} & \multicolumn{1}{c}{32.9}   & \multicolumn{1}{c}{43.8}  & \multicolumn{1}{c}{49.2} & \multicolumn{1}{c}{27.2}  & \multicolumn{1}{c}{45.1} & \multicolumn{1}{c}{36.4}  & \multicolumn{1}{c}{30.3}   & \multicolumn{1}{c}{34.6}    & \multicolumn{1}{c}{37.4} \\
PSA-ART~\cite{zheng2020cross}    & \multicolumn{1}{c}{34}     & \multicolumn{1}{c}{46.9}  & \multicolumn{1}{c}{52.1} & \multicolumn{1}{c}{\textbf{30.8}}  & \multicolumn{1}{c}{43.2} & \multicolumn{1}{c}{29.9}  & \multicolumn{1}{c}{34.7}   & \multicolumn{1}{c}{37.4}    & \multicolumn{1}{c}{38.6} \\
ATF~\cite{he2020domain}        & \multicolumn{1}{c}{34.6}   & \multicolumn{1}{c}{47}    & \multicolumn{1}{c}{50}   & \multicolumn{1}{c}{23.7}  & \multicolumn{1}{c}{43.3} & \multicolumn{1}{c}{38.7}  & \multicolumn{1}{c}{33.4}   & \multicolumn{1}{c}{\textbf{38.8}}    & \multicolumn{1}{c}{38.7} \\
FBC~\cite{yang2020unsupervised}       & \multicolumn{1}{c}{31.5}   & \multicolumn{1}{c}{46}    & \multicolumn{1}{c}{44.3} & \multicolumn{1}{c}{25.9}  & \multicolumn{1}{c}{40.6} & \multicolumn{1}{c}{\textbf{39.7}}  & \multicolumn{1}{c}{29}     & \multicolumn{1}{c}{36.4}    & \multicolumn{1}{c}{36.7} \\
LGAAD~\cite{zhang2021local}      & \multicolumn{1}{c}{33.3}   & \multicolumn{1}{c}{45.8}  & \multicolumn{1}{c}{44.6} & \multicolumn{1}{c}{25.7}  & \multicolumn{1}{c}{39.4} & \multicolumn{1}{c}{30}    & \multicolumn{1}{c}{32.5}   & \multicolumn{1}{c}{37.2}    & \multicolumn{1}{c}{36.1} \\
ILMDA~\cite{wei2020incremental}     & \multicolumn{1}{c}{30.4}   & \multicolumn{1}{c}{45.5}  & \multicolumn{1}{c}{44.2} & \multicolumn{1}{c}{24.8}  & \multicolumn{1}{c}{50.4} & \multicolumn{1}{c}{32.3}  & \multicolumn{1}{c}{28.4}   & \multicolumn{1}{c}{30.9}    & \multicolumn{1}{c}{35.9} \\
SL-DA~\cite{xiong2021domain}     & \multicolumn{1}{c}{\textbf{38.4}}   & \multicolumn{1}{c}{\textbf{47.7}}  & \multicolumn{1}{c}{\textbf{52.7}} & \multicolumn{1}{c}{29.5}  & \multicolumn{1}{c}{44.9} & \multicolumn{1}{c}{31.6}  & \multicolumn{1}{c}{34.3}   & \multicolumn{1}{c}{38.7}    & \multicolumn{1}{c}{39.7} \\
Ours               & \multicolumn{1}{c}{32.9}   & \multicolumn{1}{c}{46.8}  & \multicolumn{1}{c}{49.8} & \multicolumn{1}{c}{30.4}  & \multicolumn{1}{c}{\textbf{54.4}} & \multicolumn{1}{c}{33}    & \multicolumn{1}{c}{\textbf{35.9}}   & \multicolumn{1}{c}{36.1}    & \multicolumn{1}{c}{\textbf{39.9}} \\ 
\toprule[2pt]
\end{tabular}
\label{clear2foggy}
\end{table}

\subsection{Similar Experiments}
\label{S:5.2}
\textbf{Weather Changes.} Weather changes are the most common phenomenon in autonomous driving, and they are one of the most critical safety factors. To investigate weather adaptation from clear weather to a foggy environment, we use Cityscapes as the source domain and Foggy Cityscapes as the target domain. The training sets from two datasets are used as the source and target images for training in this experiment. In addition, 500 images from the foggy Cityscapes validation sets are used for testing.

As shown in Table~\ref{clear2foggy}, when compared to the Baseline method SW-faster \cite{saito2019strong}, the mAP of our methods improves by 5.1\%, and when compared to the SW-ICR-CCR model of the same series \cite{xu2020exploring}, our method improves by 2.5\%. The results show that our method has absolute improvements in "bus" and "mcycle" with the AP evaluation criteria when compared to state-of-the-art methods such as PSA-ART \cite{zheng2020cross}, SL-DA \cite{xiong2021domain}, LGAAD \cite{zhang2021local}, and so on. Furthermore, the experimental results show that our method effectively reduces domain shifts.

\textbf{Scene Changes.} Cars, as we all know, frequently travel to and from various urban settings. However, because different cities have different style layouts, there is bound to be a domain bias in autonomous driving. To investigate scene adaptation, we use Cityscapes as the source domain and the "daytime" sub-dataset of BDD100K as the target domain. And we use the same classes as Cityscapes, with the exception of "train," to evaluate the AP and mAP of all models on the BDD100K testing set.

Our method achieves the best performance in all categories and the best detection performance on BDD100K, as shown in Table~\ref{city2BDD}. When compared to the SW-faster \cite{saito2019strong}, our method's mAP improves by 5.3 percent, indicating that our method is effective at closing the domain gap and achieving better distributional alignment between the source and target domains.

\begin{table}[]
\centering\caption{ Scene Changes: Results on the daytime subset of BDD100k, all the experiments use the Cityscapes train set as the source domain and the BDD100K train set as the target domain.}
\begin{tabular}{l|ccccccccc}
\toprule[2pt]
                     & person & rider & car  & truck & bus  & train & mcycle & bicycle & mAP  \\
\hline
Faster R-CNN~\cite{ren2016faster} & 26.9   & 22.1  & 44.7 & 17.4  & 16.7 & -      & 17.1   & 18.8    & 23.4 \\
Baseline: SW-faster~\cite{saito2019strong}    & 30.2   & 29.5  & 45.7 & 15.2  & 18.4 & -      & 17.1   & 21.2    & 25.3 \\
DA-faster~\cite{chen2018domain}    & 29.4   & 26.5  & 44.6 & 14.3  & 16.8 & -      & 15.8   & 20.6    & 24   \\
DA-ICR-CCR~\cite{xu2020exploring}   & 29.3   & 28.4  & 45.3 & 17.5  & 17.1 & -      & 16.8   & 22.7    & 25.3 \\

SW-ICR-CCR~\cite{xu2020exploring}   & 31.4   & 31.3  & 46.3 & 19.5  & 18.9 & -      & 17.3   & 23.8    & 26.9 \\
Ours                 & \textbf{32.9}  & \textbf{31.8}& \textbf{47.5}& \textbf{20.8}  & \textbf{21.9} & - & \textbf{19}  & \textbf{27.1} & \textbf{28.7} \\ 
\toprule[2pt]
\end{tabular}
\label{city2BDD}
\end{table}

\begin{table}[t]
\centering
\caption{Real Image to Clipping Image. Clipart1k testing results can be found here. All methods make use of Pascal VOC as the source domain and Clipart1k as the target domain.}
\resizebox{0.9\linewidth}{!}{
\begin{tabular}{p{3.5cm}|cccccccc}
\toprule[2pt]
\multirow{3}{*}{}                                                              & aero  & bike  & bird  & boat  & bottle & bus   & car    & \multirow{3}{*}{mAP}  \\
                                                                               & chair & cow   & table & dog   & hourse & mbike & person &                       \\
                                                                               & plant & sheep & sofa  & train & tv     & cat   & -      &                       \\ \hline
\multirow{3}{*}{\begin{tabular}[c]{@{}l@{}}Faster R-CNN~\cite{ren2016faster} \end{tabular}}        & 21.9  & 42.2  & 22.9  & 19    & 30.8   & 43.1  & 28.9   & \multirow{3}{*}{27}   \\
                                                                               & 27.4  & 18.1  & 13.5  & 10.3  & 25     & 50.7  & 39     &                       \\
                                                                               & 37.4  & 6.9   & 18.1  & 39.2  & 34.9   & 10.7  & -      &                       \\
\hline
\multirow{3}{*}{\begin{tabular}[c]{@{}l@{}}Baseline:\\ SW-faster~\cite{saito2019strong}\end{tabular}} & 29.2  & 53.1  & 30.2  & 24.4  & 41.4   & 52.5  & 34.6   & \multirow{3}{*}{36.8} \\
                                                                               & 36.3  & 43.5  & 17.6  & 16.6  & 33.4   & \textbf{78.1}  & 59.1   &                       \\
                                                                               & 42.1  & 15.8  & 24.9  & 45.5  & 43.7   & 14    & -      &                       \\
\hline
\multirow{3}{*}{\begin{tabular}[c]{@{}l@{}}SW-ICR-CCR~\cite{xu2020exploring}\end{tabular}}         & 28.7  & 55.3  & 31.8  & 26    & 40.1   & 63.6  & 36.6   & \multirow{3}{*}{38.3} \\
                                                                               & 38.7  & 49.3  & 17.6  & 14.1  & 33.3   & 74.3  & 61.3   &                       \\
                                                                               & 46.3  & 22.3  & 24.3  & 49.1  & 44.3   & 9.4   & -      &                       \\
\hline
\multirow{3}{*}{DA-faster~\cite{chen2018domain}}                                                     & 38    & 47.5  & 27.7  & 24.8  & 41.3   & 41.2  & 38.2   & \multirow{3}{*}{34.7} \\
                                                                               & 36.8  & 39.7  & 19.6  & 12.7  & 31.9   & 47.8  & 55.6   &                       \\
                                                                               & 46.3  & 12.1  & 25.6  & 51.1  & 45.5   & 11.4  & -      &                       \\
\hline
\multirow{3}{*}{\begin{tabular}[c]{@{}l@{}}DA-ICR-CCR~\cite{xu2020exploring}\end{tabular}}         & 30.2  & 57    & 30.6  & 26.2  & 38     & 57.1  & 36.1   & \multirow{3}{*}{36.7} \\
                                                                               & 36.4  & 44.8  & 18.2  & 14.6  & 30     & 56.7  & 56.6   &                       \\
                                                                               & 45.9  & 17.8  & 25.3  & 50.5  & 48.5   & 12.7  & -      &                       \\
\hline
\multirow{3}{*}{UaDaN \cite{guan2021uncertainty}}                                                         & 35    & \textbf{72.7}  & \textbf{41}    & 24.4  & 21.3   & 69.8  & \textbf{53.5}   & \multirow{3}{*}{40.2} \\
                                                                               & 34.2  & \textbf{61.2}  & \textbf{31}    & \textbf{29.5}  & \textbf{47.9}   & 63.6  & 62.2   &                       \\
                                                                               & \textbf{61.3}  & 13.9  & 7.6   & 48.6  & 23.9   & 2.3   & -      &                       \\
\hline
\multirow{3}{*}{FBC~\cite{yang2020unsupervised}}                                                           & \textbf{43.9}  & 64.4  & 28.9  & 26.3  & 39.4   & 58.9  & 36.7   & \multirow{3}{*}{38.5} \\
                                                                               & \textbf{46.2}  & 39.2  & 11    & 11    & 31.1   & 77.1  & 48.1   &                       \\
                                                                               & 36.1  & 17.8  & \textbf{35.2}  & \textbf{52.6}  & \textbf{50.5}   & \textbf{14.8}  & -      &                       \\
\hline
\multirow{3}{*}{Ours}                                                          & 34.4  & 57.5  & 34.3  & \textbf{27.9}  & \textbf{41.6}   & \textbf{70.8}  & 36.3   & \multirow{3}{*}{\textbf{40.7}} \\
                                                                               & 42.4  & 61.1  & 11    & 22.6  & 26.7   & 72.4  & \textbf{63.3}   &                       \\
                                                                               & 46.7  & \textbf{23}    & 29.8  & 52.1  & 48.2   & 11.6  & -      &                       \\ \toprule[2pt]
\end{tabular}
}
\label{p2c}
\end{table}

\begin{table}[]
\centering
\caption{AP on adpatation from PASCAL VOC to WaterColor2K (\%), which is tested on WaterColor2K. }
\begin{tabular}{l|ccccccc}
\toprule[2pt]
\multicolumn{1}{c}{} & bike & bird & car  & cat  & dog  & person & mAP  \\ \hline
Faster R-CNN~\cite{ren2016faster} & 66.7 & 43.5 & 41   & 26   & 22.9 & 58.9   & 43.2 \\
Baseline: SW-faster~\cite{saito2019strong}   & 82.3 & 55.9 & 46.5 & 32.7 & 35.5 & 66.7   & 53.3 \\
DA-faster~\cite{chen2018domain}    & 75.2 & 40.6 & 48   & 31.5 & 20.6 & 60     & 46   \\
FBC~\cite{yang2020unsupervised}         & \textbf{90.1} & 49.7 & 44.1 & 41.1 & 34.6 & \textbf{70.3}   & 55   \\
ATF~\cite{he2020domain}          & 78.8 & \textbf{59.9} & 47.9 & 41   & 34.8 & 66.9   & 54.9 \\
ILMDA~\cite{wei2020incremental}       & 82   & 52.8 & 49.5 & 36   & 31.5 & 65.1   & 52.8 \\
LGAAD~\cite{zhang2021local}        & 89.5 & 52.6 & 48.9 & 35.1 & 37.6 & 64.9   & 54.8 \\
Ours                 & 83.9 & 52.7 & \textbf{50.3} & \textbf{41.7} & \textbf{39.2} & 63.7   & \textbf{55.3} \\ \toprule[2pt]
\end{tabular}
\label{p2w}
\end{table}

\subsection{Dissimilar Experiments}
\label{S:5.3}
There are numerous datasets with significant domain biases in real-world applications. Experiments are carried out to validate the efficacy of our method in these situations. In the dissimilar domain experiments, ResNet-101 \cite{he2016res101} is used as the backbone. In the ResNet-101 \cite{he2016res101}, the first two modules' parameters are fixed, while the last three modules are used for domain adaptation.

\textbf{Real Image to Clipping Image.} PASCAL VOC \cite{cityscapes} is a well-known real-world object detection dataset, and Clipart1k \cite{clipart1k} is a clip dataset containing 1000 virtual images. They share the same 20 categories, but there is a significant domain shift between the two data sets. The source domain in this case is PASCAL VOC, and the target domain is all images in Clipart1k. And we put our method to the test on Clipart1k, using all 1000 images as a validation set.

The results are shown in the table. \ref{p2c}, our model improves by 3.9\% and 2.4\% over the baseline SW-faster and the same family of models SW-ICR-CCR, respectively. At the same time, our model outperforms other models. It demonstrates that our method can reduce domain bias more effectively than other methods for the dissimilar domain.

\textbf{Real Image to Watercolor Image.} We used the Pascal VOC dataset as the source domain and Watercolor2K as the target domain after configuring SW-faster. Watercolor2K has a total of 2K images and six categories that are compatible with PASCAL VOC. The 1K images in the Watercolor2K dataset were used for training, while the other 1K images were used as validation sets for testing.

Our model achieves the best detection results in the "car," "cat," and "dog" categories, as shown in the Table~\ref{p2w}. Furthermore, the mAP of our method is 2\% higher than that of the SW-faster. It means that in dissimilar domain adaptation tasks, our method can reduce the domain gap better than the baseline. Simultaneously, our model outperforms other models. It also shows that our method is capable of more effectively transferring knowledge from the source domain to the target domain.

\subsection{Ablation Study}
\label{S:5.4}
 \begin{figure}[t]
\centering
\includegraphics[width=0.7\linewidth]{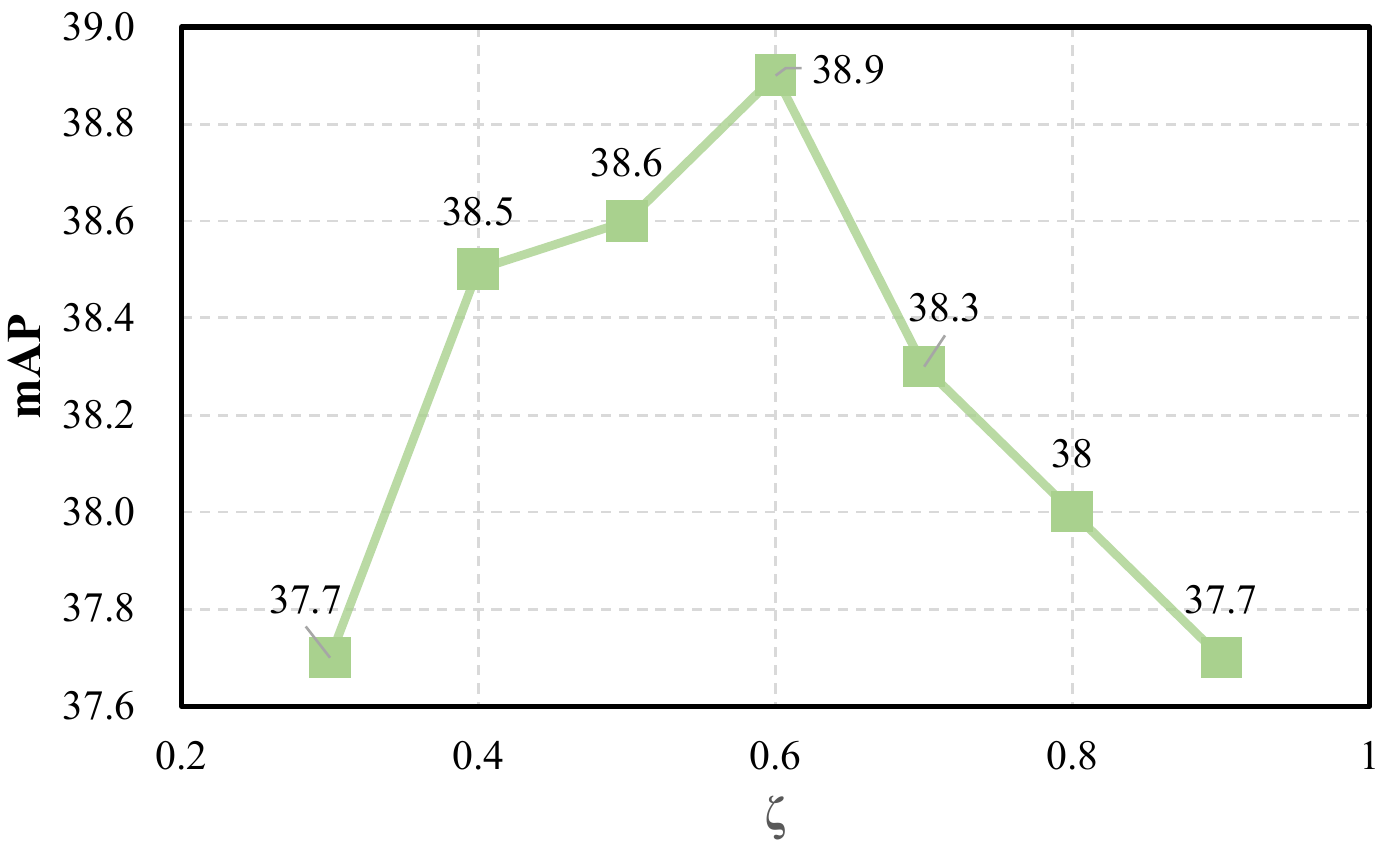}
\caption{The result of testing on Foggy Cityscapes with different semantic calibration hyperparameter $\zeta$.}
\label{S-CA}
\end{figure}

 \begin{figure}[t]
\centering
\includegraphics[width=0.7\linewidth]{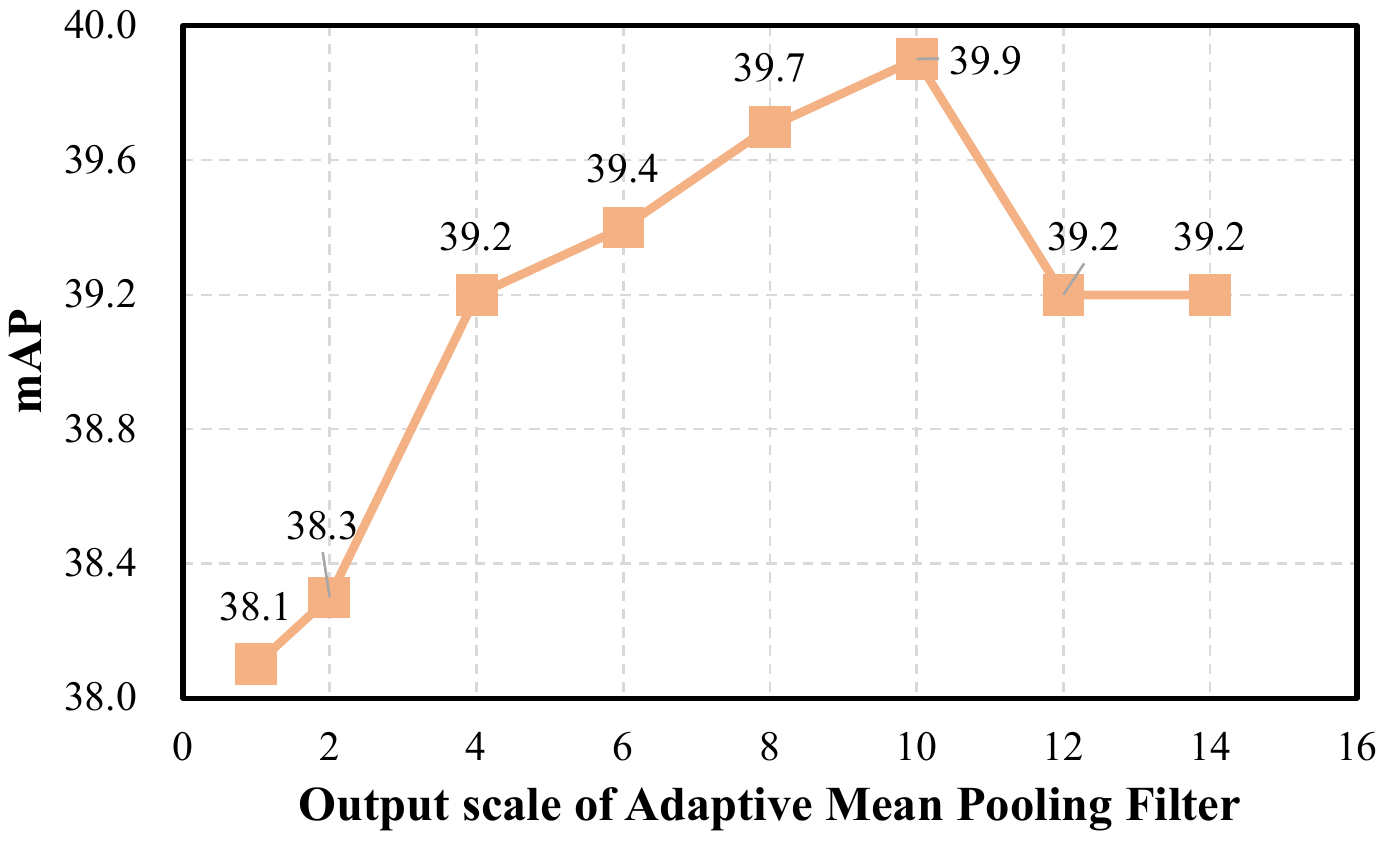}
\caption{The result of testing on Foggy Cityscapes with different output scales of the adaptive mean pooling filter. And the $H_a$ is the same with $W_a$}.
\label{S-Pool}
\end{figure}

\textbf{Semantic Calibration Hyperparameter:} The semantic calibration hyperparameter $\zeta$ is critical for the semantic prediction modules, as it determines the types of categories in the receptive field. In these experiments, Cityscapes \cite{cityscapes} are the source domain, Foggy Cityscapes \cite{foggy-cityscapes} are the target domain, and the model does not include Semantic-based Attention Maps and Semantic Consistency Regularization. As shown in Figure~\ref{S-CA}, when $\zeta$ is 0.6, the mAP is the best at $38.9\%$.

\textbf{Semantic Consistency Regularization Hyperparameter:} The anti-interference ability of the model is affected by the output scales of the adaptive mean pooling filter in the semantic consistency regularization component. Here, the Cityscapes \cite{cityscapes} is the source domain, the Foggy Cityscapes \cite{foggy-cityscapes} is the target domain, and the model is the same as in previous similar domain experiments. When the output scale of the adaptive mean pooling filter is $10*10$, the mAP of the model is the best at $39.9$, as shown in Figure~\ref{S-Pool}. And, when the output scale of the adaptive mean pooling filter is $1*1$, the mAP of the model is $38.1$, which means the singular values seriously influence the results of the model compared with the $10*10$.

\textbf{Component Testing:} 
The domain adaptive detection model in this paper is mainly composed of Semantic Prediction Modules(SPM), Semantic Bridging Component(SBC), Semantic-based Attention Map(SAM), and Semantic Consistency Regularization(SCR). This paper increases the components one by one for experimental to explore the effectiveness of each component. Here, the Cityscapes \cite{cityscapes} are the source domain, Foggy Cityscapes \cite{foggy-cityscapes} are the target domain, and the models are the same as the similar experiments, and all the hyperparameters are the best value. 

The results as shown in Table~\ref{C-S}. Here, the SW-faster model is a baseline. The experimental results demonstrate that all the components in this paper can improve the performance of the model. And compared with the baseline, if the SPM and SBC is added to SW-faster, the performance of the model is increased by 4.1\%, it demonstrates that SPM and SBC can achieve distributional alignment of single-class and mixed-class features of the model, thereby reducing negative transfer. 

\begin{table}[]
\centering
\caption{The ablation study of different components, and the results are testing on Foggy Cityscapes. }
\resizebox{\linewidth}{!}{
\begin{tabular}{c|ccccc|ccccccccc} 
\toprule[2pt]
Method  & MDA & SPM & SBC & ASM & SCR & person & rider & car  & truck & bus  & train & mcycle & bicycle & mAP   \\ 
\hline
SW-faster~\cite{saito2019strong} &     &     &     &     &     & 32.3   & 42.2  & 47.3 & 23.7  & 41.3 & 27.8  & 28.3   & 35.4    & 34.8  \\
\hline
\multirow{5}{*}{Ours}      & $\surd$   &     &     &     &     & 32.7   & 43.6  & 49.7 & 25.5  & 45   & 37.8  & 29.9   & 31.7    & 37    \\
          & $\surd$      & $\surd$     &     &     &     & 33.3   & 46.2  & 49.5 & 32.2  & 48.3 & 28.9  & 32.4   & 35.3    & 38.3  \\
          & $\surd$     & $\surd$      & $\surd$      &     &     & 33.2   & 45.8  & 50.7 & 29.7  & 49.6 & 32.6  & 34.4   & 35.2    & 38.9  \\
          &$\surd$      & $\surd$      & $\surd$      &$\surd$     &     & 33.5   & 44.5  & 50   & 28.5  & 51.2 & 38.1  & 32     & 36      & 39.2  \\
          & $\surd$     & $\surd$     & $\surd$      &$\surd$      & $\surd$      & 32.9   & 46.8  & 49.8 & 30.4  & 54.4 & 33    & 35.9   & 36.1    & 39.9  \\
\toprule[2pt]
\end{tabular}
}
\label{C-S}
\end{table}


\section{Conclusion} 
\label{S:6}
In current object detection tasks, $H-divergence$ lacks the distribution distance measure of mixed-class features. The $MCH-divergence$ method is proposed in this paper for designing domain adaptive object detection models. To extract all classes in each feature, semantic prediction networks and semantic bridge components are proposed. The semantic features in the semantic prediction networks are then spliced into the original features via semantic bridge components, which can strengthen the semantics of each feature and achieve the separability of single-class and mixed-class features. Simultaneously, this paper proposes a semantic attention loss based on semantic prediction network output to solve the imbalance of positive and negative samples, which increases the weight of a few samples in different feature layers to achieve a certain degree of sample balance. A category of the semantic consistency loss constraint feature is designed with the goal of ensuring that the semantics of the global prediction of the source domain data is consistent with the semantics of the intermediate prediction. Experiments with different domain differences show that the method described in this paper can effectively improve model performance.

\section{Acknowledgements} 
\label{S:7}
This research did not receive any specific grant from funding agencies in the public, commercial, or not-for-profit sectors. 

\bibliographystyle{elsarticle-num-names}
\bibliography{sample.bib}
\end{document}